%% file: main.tex
\documentclass[11pt,a4paper]{article}
\usepackage[hyperref]{emnlp2020}
\usepackage{times}
\usepackage{latexsym}
\usepackage{amsmath}
\usepackage{amssymb}
\usepackage[utf8]{inputenc}

\usepackage[shortlabels]{enumitem}
\setlist[enumerate]{itemsep=0pt}

\usepackage{url}

\aclfinalcopy 

\title{Internal and external pressures on language emergence: \\ least effort, object constancy and frequency}

\author{Diana  Rodr\'{i}guez Luna \\
  University of Amsterdam \\
  {\tt diana.rodrigzluna@gmail.com} \\\And
  Edoardo Maria Ponti \\
  LTL, University of Cambridge \\
  {\tt ep490@cam.ac.uk }
  \\\AND
  Dieuwke Hupkes\thanks{~ Shared senior authorship} \\
  ILLC, University of Amsterdam \\
  {\tt d.hupkes@uva.nl}
  \\\And
  Elia Bruni$^*$ \\
  Universitat Pompeu Fabra \\
  {\tt elia.bruni@gmail.com} \\}

\date{}

\usepackage{enumitem}
\usepackage{graphicx}
\usepackage{todonotes}
\usepackage{subfig}

\graphicspath{ {images/} }

\begin{document}
\maketitle

\input{abstract}

\input{introduction}

\input{motivation}

\input{setup}

\input{variants}

\input{experiments}

\input{results}

\input{conclusion}

\input{acknowledgements}

\bibliography{main}
\bibliographystyle{acl_natbib}

\clearpage
\input{appendix}

\end{document}

%% file: abstract.tex
\begin{abstract}

In previous work, artificial agents were shown to achieve almost perfect accuracy in referential games where they have to communicate to identify images. 
Nevertheless, the resulting communication protocols rarely display salient features of natural languages, such as compositionality.
In this paper, we propose some realistic sources of pressure on communication that avert this outcome.
More specifically, we formalise the principle of \textit{least effort} through an auxiliary objective. Moreover, we explore several game variants, inspired by the principle of \textit{object constancy}, in which we alter the frequency, position, and luminosity of the objects in the images. We perform an extensive analysis on their effect through compositionality metrics, diagnostic classifiers, and zero-shot evaluation.
Our findings reveal that the proposed sources of pressure result in emerging languages with less redundancy, more focus on high-level conceptual information, and better abilities of generalisation.
Overall, our contributions reduce the gap between emergent and natural languages.
\end{abstract}

%% file: introduction.tex
\section{Introduction}
\label{sec:introduction}

One of the key requirements for a machine to be intelligent is its ability to communicate in natural language \cite{Mikolov_2018}.
While supervised approaches with labelled texts have recently achieved unprecedented performances in several applications \citep[\textit{inter alia}]{chen2017survey}, they still neglect fundamental components of natural communication, such as the speakers' intention and the function of their utterances \cite{clark1996using}.

This functional aspect of language instead is captured by multi-agent games \cite{kirby2002natural}, in which agents have to communicate about some shared input space (e.g.\ images). Agents usually manage to communicate with success, measured in terms of \emph{task accuracy} \citep[\textit{inter alia}]{mordatch2017emergence,choi2018compositional}, if the setting is fully cooperative \citep{cao2018emergent}. 
However, several studies have shown that the emerged languages rarely display features inherent to natural languages, such as compositionality of meaning\footnote{Compositionality in the context of neural networks is a complex topic \citep[see, e.g.][]{hupkes2020compositionality}. Here we assume the interpretation of \citet{frege1892}: The meaning of a complex expression is determined by its structure and the meanings of its constituents.} and generalisation to novel objects \citep[e.g.][]{kottur2017natural,vanderwal2020grammar}. For instance, agents might develop protocols to refer to specific pixel values, rather than concept-level information \cite{bouchacourt2018agents}.

Referential games are a perfect controlled environment to study how \textit{sources of pressure} on the agents
affect the `naturalness' of emergent languages. Previous work has proposed to limit the memory of neural agents across turns of dialogue \citep{kottur2017natural} or to soft-constrain the active vocabulary size \citep{mordatch2017emergence}. However, these constraints seem at odds with the capacity of human memory. In this work, we propose a set of yet unexplored but more realistic sources of pressure, either internal to the agents or external, pertaining to the input space.

An internal source of pressure, inspired by the principle of least effort \citep[see \S~\ref{ssec:leasteffort}]{zipf1935psycho,haiman1983iconic}, compels the agents to keep the length of sentences to the bare minimum. We implement this pressure through an auxiliary loss that incentivises the generation of the end-of-sentence token as early as possible. Several external pressures instead are implemented as game variants, where we control for the frequency, the position, and the illumination of objects in images. These game variants are again motivated by principles governing human perception, such as object constancy \citep[see \S~\ref{ssec:objconst}]{lorenz1977behind,gillam2000perceptual}.


Our results demonstrate that the internal pressure efficiently compresses the sentence lengths and the vocabulary size without loss of accuracy. Moreover, based on established metrics of compositionality \citep{choi2018compositional,lazaridou2018emergence,bouchacourt2018agents} and zero-shot evaluation, we show that agents with pressure towards object constancy achieve the highest scores. Finally, diagnostic classification reveals how the external pressures make agents sensitive to higher-level object properties.

In general, we offer a series of contributions. In addition to a novel model objective and game variants, we establish a methodology to adapt the communication hyper-parameters automatically. Moreover, we draw connections to principles of human cognition, thus aligning the multi-agent game to hypotheses on natural language evolution \citep{nowak1999evolution}. 
We elaborate on such principles in \S~\ref{sec:related_work}. In \S~\ref{sec:approach}, we outline the basic setup for the referential game and the dataset. The auxiliary loss and game variants that operationalise the cognitive principles are described in \S~\ref{sec:game_variants}. We discuss the metrics for evaluation in \S~\ref{sec:experiments} and provide the results in \S~\ref{sec:results}. The main conclusions to our work are summarised in \S~\ref{sec:conclusion}.

%% file: motivation.tex
\section{Motivation}
\label{sec:related_work}
The proposed sources of pressure on emergent languages are motivated on the basis of general principles of human communication and perception. In this section, we outline the principles of least effort, object constancy, and object frequency.

\subsection{Least effort}
\label{ssec:leasteffort}
While human speakers try to maximise the distinctiveness of the information conveyed, they also minimise the effort involved. A version of this principle was originally formulated by \citet[p. 29]{zipf1935psycho} -- who pointed out the correlation between word frequency rank and word length -- and was later generalised to every reduction of linguistic expressions by \citet{haiman1983iconic} under the name `Principle of Economy.' This principle is also reminiscent of the  maxim of quantity in pragmatics \citep{grice1975logic,levinson2000presumptive}, which requires to give no more information than needed. 

As this principle is a key factor in explaining the variation of natural languages \citep{haiman1983iconic}, it posits a realistic constraint on language emergence. Moreover, our operationalisation of the principle allows the model to determine automatically the maximum length of sequences and the number of symbols in its vocabulary. In doing so, the complexity of the emergent language is gauged according to the data and task at hand. This has the methodological advantage of not requiring to preset these hyper-parameters arbitrarily or performing grid search on task accuracy (which rarely corresponds to natural language properties).

\subsection{Subjective constancy}
\label{ssec:objconst}

Reality as it is immediately sensed is shapeless and ever-changing. However, animals evolved to various degrees the ability to perceive a reality of objects, namely constant and discrete entities lying behind the tangle of sensation \citep{lorenz1977behind}. 
The same ability is connected with abstraction: over repeated impressions, animals learn to neglect what is contingent (due to the environment or their internal disposition), and group instances of objects with recurring patterns into the same conceptual class \citep{gillam2000perceptual}. 

Object constancy involves several different and independent mechanisms, regarding, among others: i) the \textbf{colour} of the object, under different light conditions \citep{holst1957aktive}; ii) the \textbf{position} of the object, under different perspectives \citep{von1970verhaltensphysiologie}. For instance, bees have to identify flowers by their colour independently from the time of the day (red of dusk or gold of dawn). In our implementation, we manipulate the images in such a way that agents are exposed to the same object with different position or luminosity. 
As a consequence, we expect the agents to acquire some sort of constancy mechanisms.

\subsection{Object frequency}
\label{ssec:objfreq}
The distribution of objects and features in the real world are highly non-uniform. Agents encounter objects in the environment with different frequencies. Furthermore, the degree of association between features and objects can vary: for instance, berries evoke the colour blue less vividly than the sky. Frequency facilitates the correct classification of object instances \citep{nosofsky1988similarity}. Moreover, \citet{medin1978context} have shown that more frequent stimuli lead to an increasing perceptual differentiation in the region of their features. As a consequence, agents are imprinted with respect to specific features rather than the stimulus as a whole, and stimuli become decomposable into their `building blocks' \citep{schyns1998development}. Recently, \citet{hendrickson2019cross} have also shown how a Zipfian distribution of words and referents can accelerate word meaning acquisition compared to a uniform one.

%% file: setup.tex
\section{Setup}
\label{sec:approach}

We study language emergence in the context of task-oriented multi-agent games.
In the current section, we present our baseline setup (\S~\ref{sec:game_description}) and the dataset that we use (\S~\ref{sec:data}).

\subsection{Game definition}
\label{sec:game_description}

In the game we study, two agents play a referential game. One agent, the Sender, has to describe an image; the other agent, the Receiver, has to pick the correct image out of a line-up of confounders. We follow the setup of \citet{havrylov2017emergence}:\vspace{-3mm}
\begin{enumerate}
\item There are $N$ images represented by z-dimensional feature vectors $f_n=\{i_1,...,i_z\}$. A target image $t$ is sampled and shown to the Sender.
\item The Sender generates a message $m$ with a maximum length $L$ that consists of a sequence of words $\{w_i, \dots, w_{\leq L}\}$ from a vocabulary of size $|V|$.
\item The Receiver uses $m$ to identify $t$ in a set of images that contains $k$ distracting images and $t$ in random order.
\end{enumerate}
\vspace{-3mm}

We implement both the Sender and Receiver agents as LSTM networks. Unless otherwise specified, we follow again the training procedures and error definitions of \citet{havrylov2017emergence}.\footnote{For brevity, we omit these details from the full paper, but report them in Appendix~A.}
A scheme of this setup is shown in Figure~\ref{fig:modelDiagram}.

\begin{figure*}
\center
\includegraphics[width=1.\textwidth, trim=0mm 3mm 0mm 2mm, clip]{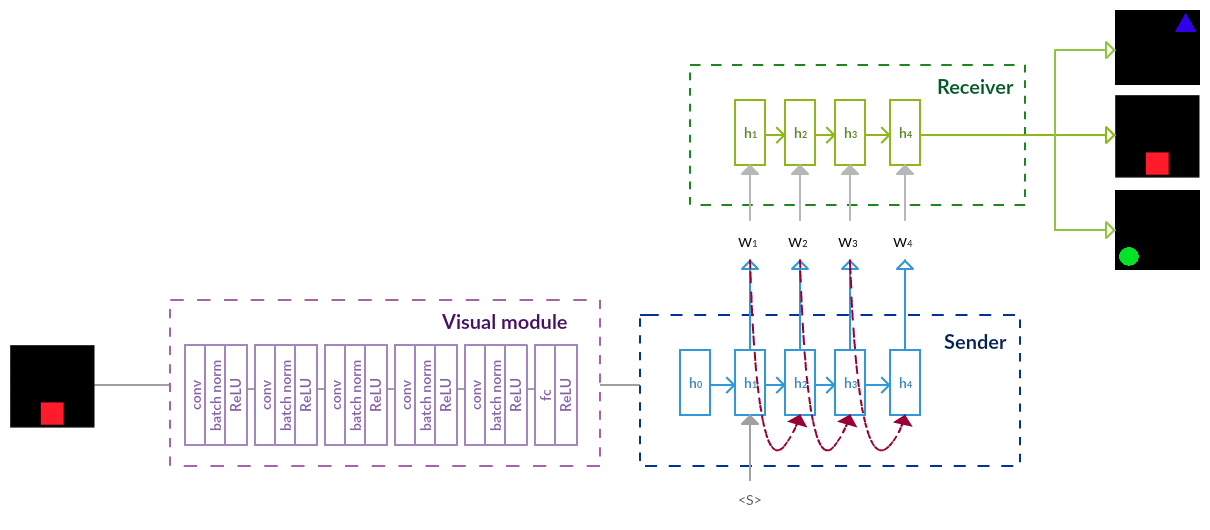}
\caption{Architecture overview of the Sender and Receiver. The visual module shows the CNN architecture used for extracting features from the input.}
\label{fig:modelDiagram}
\end{figure*}

\subsection{Data}\label{sec:data}

We use images from a modified version of the SHAPES dataset \citep{Andreas_2016}. This dataset consists of 30 x 30 pixel images. Each image contains exactly one 2D object, which is characterised by a shape (circle, square, triangle), a colour (red, green, blue), and a size (small, big). 
The objects are positioned in a logical grid of three rows and three columns. In the baseline setting, 
we sample both images and distractors uniformly from the images in this space. In the next section, we introduce three alternative versions of the game, in which images are selected following more naturalistic procedures.\footnote{While we work on synthetic data, the same expedients can be easily applied to natural datasets like COCO \citep{lin2014microsoft}.}

%% file: variants.tex
\section{Formalisation of pressures}
\label{sec:game_variants}

As the core contribution of this paper, we propose a series of changes to the baseline setup in order to incorporate model internal and external pressures related to concept learning.
In the current section, we describe these alternative setups.

\subsection{Least effort pressure}


Arguably, communicative success is not the only factor that comes into play in natural interactions. 
In fact, agents should also abide by the principle of least effort.
We formalise this idea with a \textit{vocabulary loss}, that encourages the agents to use shorter messages and fewer words. For each time step \textit{t}, the logits $s$ of the Sender are squashed into a predictive probability distribution over the vocabulary $p(V \mid \cdot)$ through softmax. Then we estimate its cross-entropy with a distribution $q(V)$ where the whole probability mass is assigned to the generated word $w \triangleq \max_{w \in V} p(V \mid \cdot)$. The formula can be written as:

\begin{equation*}
\begin{aligned}
\mathcal{L}_v= \sum_{1 \leq t \leq L} - \log \frac{\textrm{exp}(s_w^{(t)})}{\sum_{j} \textrm{exp}(s_j^{(t)} )}
\end{aligned}
\end{equation*}
This vocabulary loss encourages high-confidence predictions, since it is minimised when $p(V=w\mid\cdot)=1$. As a side effect, the model is pushed towards preferring a subset of the vocabulary. Moreover, as the end of sequence symbol \textless$S$\textgreater~is part of the vocabulary, the loss is also implicitly favouring shorter sentences.\footnote{We experimented with adding an additional parameter $\alpha$ to explicitly scale the counts of \textless$S$\textgreater and modulate its emission. However, we found the best value of $\alpha$ to be 1.}
This auxiliary loss is added to the main loss of the system, with a weighting factor $\lambda$,\footnote{We set the value of the weighting factor $\lambda$ as a hyperparameter through grid search (over 0.1, 0.2 and 0.4), obtaining the best results for 0.1.} and minimised during optimisation.


\subsection{Location invariance} 
\label{ssec:locinv}
Among the cognitive mechanisms governing object constancy and abstraction, a key role is played by location invariance. This mechanism allows animals to conceive objects as identical even when they move, and the object reflection on their retina has shifted (see \S\ \ref{ssec:objconst}). 
We formalise the pressure to develop location invariant concepts by introducing a mismatch between the exact object instance shown to the Sender and Receiver.
More precisely, when the Sender is shown a target image $t$, characterised by the quintuple (color, size, shape, horizontal position, vertical position) -- the target image $t'$ of the Receiver contains an object with the same shape, colour, and size, but a different position. 
We hypothesise that this setup will encourage the emergence of general purpose symbols for colour, shape, and size, since agents are pressured to refer to these concepts consistently across different perspectives.


\subsection{Colour constancy}

Another object-constancy mechanism allows animals to identify objects with altered hues, when the conditions of illumination  change (see \S\ \ref{ssec:objconst}). To introduce this pressure in our game setup, we follow a similar protocol as in \S~\ref{ssec:locinv}: we encourage colour constancy by slightly perturbing the Sender's target image $t$ into the target image of the Receiver $t'$.
More specifically, the target image of Sender and Receiver are identical in all dimensions, except their overall \textit{brightness}.
Therefore, two different \textit{brightness} values $b_1$ and $b_2$ are assigned to each image, so that:
\begin{gather*}
0.2 < b_1 < 0.8 \textrm{,} \quad 0.2 < b_2 < 0.8 \textrm{,} \\ \textrm{and} \quad |b_1 - b_2| \geq 0.2
\end{gather*}

\subsection{World distribution}
Finally, we consider the effect of frequency on concept memorisation, by exposing agents to non-uniform distributions of objects (see \S\ \ref{ssec:objfreq}). 
To obtain such a non-uniform world, we skew the distributions of different shapes $p(shape)$, as well as the conditional probability distribution $p(colour|shape)$.
In particular, we sample the probabilities such that for all pairs of distinct shapes $s_1, s_2\in shapes$, it holds that:\vspace{-3mm}
\begin{equation*}
\mid (p(s_1) - p(s_2)) \mid \leq 0.2 
\end{equation*}
And for all pairs of distinct colours $c_1, c_2\in colours$ and all $s\in shapes$:\vspace{-2mm}
\begin{equation*}
\mid (p(c_1|s) - p(c_2|s)) \mid \leq 0.8 
\end{equation*}

We sample images from these new distributions. In the resulting worlds, shapes are more likely to have some colours rather than others, and some shapes are more likely than others.

%% file: experiments.tex
\section{Experiments}
\label{sec:experiments}

We now describe the procedures that we use for training and evaluation, and provide details on the (hyper)parameter settings.

\subsection{Architecture and hyperparameters}
The Sender and Receiver LSTMs have an embedding size of 256 and a hidden layer size of 512.
The number of distracting images $k$ for the Receiver is 3 for all experiments.
The initial vocabulary size $V$ of the Sender is 25 and the maximum message length $L$ is 10.
We empirically set the weighting factor $\lambda$ for the vocabulary loss to 0.1.\footnote{We found this to be the most stable value in terms of accuracy and emerging language properties.}

To pre-process the images before they are given to the agents, we use a visual module inspired by the CNN architecture of \citet{choi2018compositional}, consisting of five convolutional layers followed by batch normalisation with ReLU as activation function.
Each layer has 20 filters with a kernel size 3 and stride 2 with no padding.
This block is followed by a fully connected layer that produces 2048-dimensional feature vectors activated by a ReLU function.
The visual model is \textit{pretrained} separately for each game variant, by playing that variant of the game once, resulting in four visual modules that are specialised for a particular game variant.
The task accuracies during pretraining were all between 0.9 and 1.\footnote{For the specific results, see Appendix B.}


\subsection{Training}

In all games, we use 75k, 8k, 40k samples for the train, validation and test sets, respectively.
We train the agents using Gumbel-Softmax with a temperature of 1.2, a batch size of 128, and the Adam optimiser with an initial learning rate of 0.0001. 
We use early stopping with a patience of 30 to avoid overfitting.
We run every experiment 3 times and report the average results.

\subsection{Evaluation}
\label{sec:metrics}
In addition to monitoring game \textit{accuracy}, defined as the ratio of games in which the Receiver correctly identifies the target image, we evaluate the characteristics of the emerging communication with a range of different metrics that have been established in previous work.


\paragraph{Average message length} In order to understand to which extent agents manage to compress their communication, we keep track of the average number of tokens in the messages produced by the Sender.

\paragraph{Active symbols} The counterpart of the average message length is the \textit{active symbols} metric, which expresses how many symbol types from its vocabulary are used by the Sender.

\paragraph{Message distinctness} 
To succesfully complete the game, it may not be necessary to refer to all features of the input image.
Following \citet{choi2018compositional}, we used \textit{message distinctness} as an estimate of how much of the image features is captured in a message. 
Message distinctness is defined as the count of unique messages per batch divided by the batch size. As a reference point for this metric, we compute also the number of distinct images. Generating more messages than the reference point suggests that agents are using multiple messages to refer to the same picture. Conversely, generating fewer messages than the reference point indicates that agents use a shallower language, not covering all aspects of the image.

\paragraph{Perplexity per symbol.} As in \citet{havrylov2017emergence}, we used the perplexity per symbol metric to measure how often a symbol was used in a message to describe the same object:
$$ Ppl = \textrm{exp}(-\sum[ s^{(t)} \cdot \log(s^{(t)}) ]) $$
where $s^{(t)}$ are the vocabulary scores (given by an affine transformation of the Sender's hidden state at timestep $t$) for all symbols in the vocabulary.
A lower perplexity shows that the same symbols are consistently used to describe the same objects.

\paragraph{Topographic similarity} We study the topographic similarity (TS) between the message and input space, defined as the pairwise Pearson correlation between points in those spaces \cite{brighton2006understanding}.
As in \citet{lazaridou2018emergence}, we use this metric to measure the extent to which similar objects receive similar messages.

\paragraph{Language entropy} The language entropy $S$ denotes the variability of the number of symbols in the language.
It is given by the formula
$$S = -\sum_{w \in V}[ c_w \cdot \log(c_w) ]$$
where $c_w$ is the count of the occurrences in the produced messages of each symbol $w$ for all symbols in the vocabulary $V$. 

\paragraph{Representation Similarity Analysis} (RSA) is defined analogously to TS, but is computed on the continuous hidden representations of the Sender and Receiver \cite{kriegeskorte2008representational}.
As in \citet{bouchacourt2018agents}, we use this metric to measure the distance between two points in different embedding spaces. Sender-Receiver RSA indicates the RSA between the Sender's and Receiver's embedding spaces.
Sender-Input RSA and Receiver-Input RSA indicate the correlation between the agents and the input space.

%% file: results.tex
\section{Results}
\label{sec:results}

In this section, we provide the experimental results and discuss them critically. In \S~\ref{ssec:internal}, we establish the effect of the vocabulary penalty on the sequence lengths and vocabulary sizes in the emerging languages. 
In \S~\ref{subsec:zeroshot}, we compare the impact of different external pressures on the model's ability to generalise in \textit{zero-shot} evaluation. In \S~\ref{ssec:compmetr}, we analyse the protocols evolved in each game variant in the light of the metrics described in \S~\ref{sec:metrics}. Finally, we investigate which image features can be \textit{decoded} from the emitted messages in \S~\ref{subsec:diagnozze}.


\subsection{Least-effort pressure}\label{ssec:internal}

Maintaining a maximum vocabulary size and message length of 25 and 10, respectively, we train Sender-Receiver pairs with and without the penalty loss. Results are shown in Table~\ref{tab:avg_original_stats_overview} for the baseline setup, and Table \ref{tab:variant_games_test_overview} for the game variants.


\begin{table}[t]
\centering
\begin{tabular}{|c|c|c|c|c|c|} \hline
    & \textbf{Acc} & $\mathbf{\mu(\ell)}$ & $\mathbf{\sigma^2(\ell)}$ & $\mathbf{|\Sigma|}$ & $\mathbf{|M|}$ \\\hline
baseline & 0.99 & 11.0 & 0.0  & 20.67 & 4.81 \\
penalty  & 0.98 & 6.10 & 0.87 & 13.0 & 3.54 \\ \hline
\end{tabular}
\caption{Accuracy, average message length, variance of the message length, number of active symbols $\Sigma$, and average number of unique symbols per message $M$ on the test set when playing the baseline and penalised games. 
All values are averaged over three runs.}
\label{tab:avg_original_stats_overview}
\end{table}	

\paragraph{Baseline setup}
Based on Table~\ref{tab:avg_original_stats_overview}, 37\% fewer symbols were used in games trained with the penalty loss. 
The average length of the messages decreased in 45\%.
Additionally, the variance in the message length increased from 0 to 0.87, showing the variability of the sequence lengths needed to play the game, as opposed to always using the maximum allowed length.
Moreover, there is 26\% more symbol reuse within the sequences in the penalty case, as shown by the lower number of unique symbols per message.
In terms of accuracy, however, there is no clear difference between games with and without the vocabulary loss.
Using fewer words and shorter messages does not, at least in this case, hamper communication success. This indicates that the original models used unnecessarily many symbols. 

\paragraph{Game variants}
For the different game variants, the penalty has a similar effect on the language statistics (shown in Table \ref{tab:variant_games_test_overview}): fewer words are used, the average message length is shorter, and there is more word reuse per generated message.
The language compression is most evident in the location invariance setups, where fewer messages are required to fully describe the input space: two objects are considered identical if they share colour, shape and size, regardless of their position in the grid.
The models trained without penalty do not reflect this difference, and use the maximum message length they are allowed.

These results show that the use of the vocabulary loss gives rise to languages with symbol reuse.
It allows the model to dynamically adjust the vocabulary size and sequence lengths while still being able to successfully solve the game.
Given this positive result, we use the vocabulary penalty with a $\lambda=0.1$ in all subsequent experiments.

\begin{table*}[ht!]
\centering
\begin{tabular}{|c|c|c|c|c|c|} \hline
\textbf{Game} & \textbf{Penalty} &  \textbf{Acc} & $\mathbf{\mu(\ell)}$ & $\mathbf{|\Sigma|}$ & $\mathbf{|M|}$ \\ \hline
\textit{Location invariance} & Off & 0.91 &	11.00 & 12.33 & 2.85 \\
\textit{Colour constancy} & Off  & 0.99 & 11.00 & 21.67 & 3.25\\ 
\textit{World distribution} & Off  & 0.99 &	11.00 &	25.00 &4.38 \\ \hline
\textit{Location invariance} & On & 0.90 & 6.66 & 5.33 &2.36 \\
\textit{Colour constancy} & On & 0.99 & 7.49 & 10.0 & 2.64\\ 
\textit{World distribution} & On & 0.98 & 7.04 & 13.33 &3.54\\ \hline
\end{tabular}
\caption{Statistics for the game variant models calculated on the test set: accuracy, average message length, number of active symbols $\Sigma$, and average number of unique symbols per message $M$. Averages over 3 runs.
}
\label{tab:variant_games_test_overview}
\end{table*}

\subsection{Zero shot evaluation}
\label{subsec:zeroshot}
To assess how well the agents learned to generalise in the different setups, we run a zero-shot evaluation experiment where agents have to communicate about unseen objects.
Following the approach of \citet{choi2018compositional}, we retrain a model for each game variant, this time removing three objects from the training set images: red triangle, blue square, and green circle.
We then test these the retrained models on 40504 rounds of the game, where in each round the target is one of the held-out objects. 
The distractors are uniformly sampled from a set of objects containing both the training and held-out objects.
The prediction accuracies are reported in Table \ref{tab:zero_shot}.

All results are above chance level (0.25), which would be the average accuracy if the Receiver chose a random image every time out of the four candidates.
The highest communication success was obtained in the colour constancy (without penalty) and world distribution (with penalty) experiments.
Interestingly, the models are not similarly \textit{ranked} in the penalty and no penalty conditions, pointing to an interaction between the two different pressures that we do not yet understand.

\begin{table}[ht]
\centering
\begin{tabular}{|c|c|c|} \hline
\textbf{Game} & \textbf{Penalty} & \textbf{Acc} \\ \hline
\textit{Baseline} & Off & 0.60\\
\textit{Location invariance} & Off & 0.33\\
\textit{Colour constancy} & Off & \textbf{0.71} \\ 
\textit{World distribution} & Off & 0.46\\ 
\hline
\textit{Baseline} & On & 0.40 \\
\textit{Location invariance} & On & 0.36 \\
\textit{Colour constancy} & On & 0.33\\ 
\textit{World distribution} & On & \textbf{0.52} \\ \hline
\end{tabular}
\caption{Zero-shot accuracy on the four game variants. Average over three runs.}
\label{tab:zero_shot}
\end{table}

\subsection{Metrics}
\label{ssec:compmetr}
We report the values for the metrics outlined in \S~\ref{sec:metrics} for all game variants in Table~\ref{tab:all_games_test_metrics}.

\paragraph{Message distinctness} 

The number of distinct images (our reference point, as mentioned in Section \ref{sec:metrics}) for the baseline game, the colour constancy game, and the world distribution game, is 162 (3 shapes $\times$ 3 colours $\times$ 2 sizes $\times$ 3 rows $\times$ 3 columns).
Since this number is larger than the batch size, the expected message distinctness is 1.
The baseline model averaged a message distinctness of 0.7880, the colour constancy model 0.4921, and the world distribution model 0.8396. Thus, the world distribution game brings agents the closest to capturing the entirety of the image representation, a finding which will be further confirmed in \S~\ref{subsec:diagnozze}. 

In the location invariance experiment there are only 18 symbolically different images, since two objects are considered the same irrespective of their horizontal and vertical position.
With a batch size of 128, this gives an expected message distinctness of 18/128=0.14 per batch.
The model averaged a message distinctness of 0.2287, which indicates that the same objects are sometimes referred to with different messages (in other words, contrary to evidence, the model may still consider location to be a relevant property!).

\begin{table*}[ht!]
\centering
\begin{tabular}{|c|c|c|c|c|c|c|c|} \hline
\textbf{Game} & \textbf{Penalty} & \textbf{Ppl symbol} & \textbf{RSA S-R} & \textbf{RSA S-I} & \textbf{RSA R-I} & \textbf{Top. Sim.} & \textbf{Lang. entropy} \\ \hline
\textit{Baseline} & Off & 4.19 & 0.91 &	0.71 &	0.63 & 0.31 & 2.73 \\
\textit{Location inv} & Off & 3.11 & \textbf{0.96} & 0.69 & 0.69 &	\textbf{0.38} & \textbf{2.17} \\
\textit{Colour const} & Off & \textbf{2.18}&	0.91&	\textbf{0.72}&	\textbf{0.71}&	0.35&	2.82 \\ 
\textit{World distrb} & Off & 3.17&	0.89&	0.66&	0.62&	0.28& 3.00\\ \hline
\textit{Baseline} & On & 1.74 & 0.95 & 0.46 & 0.45 & 0.20 & 1.61 \\
\textit{Location inv} & On & 1.82 & 0.97 & \textbf{0.58} & \textbf{0.62} & \textbf{0.30} & \textbf{1.59} \\
\textit{Colour const} & On & \textbf{1.32} & \textbf{0.98} & 0.51 & 0.52 & 0.24 & 1.73 \\ 
\textit{World distrb} & On & 1.38 & 0.96 & 0.36 & 0.38 & 0.11 & 1.63 \\
\hline
\end{tabular}
    \caption{Metrics on the test set: perplexity per symbol (Ppl symbol), RSA Sender-Receiver (RSA S-R), RSA Sender-Input (RSA S-I), RSA Receiver-Input (RSA R-I), topographic similarity (Top. Sim.), and language entropy (Lang. entropy). 
    Formulas and further explanation of these metrics is given in \S~\ref{sec:metrics}.
The values for perplexity per symbol and language entropy are unbounded, all other metrics are bound between 0 and 1.
    Reported numbers are averages of three different runs per configuration.}
\label{tab:all_games_test_metrics}

\vspace{25pt}
\begin{tabular}{|c|c|c|c|c|c|c|} \hline
\textbf{Game} & \textbf{Penalty} & \textbf{Shape} & \textbf{Colour} & \textbf{Size} & \textbf{Row} & \textbf{Column}  \\ \hline
\textit{Baseline} & Off & 0.56& 0.84&	0.86	&0.98&	0.98\\
\textit{Location invariance} & Off &	0.84&1.00&	1.00&	0.33&	0.33\\
\textit{Colour constancy} & Off 	&0.54 &0.82	&0.81	&1.00	&1.00 \\ 
\textit{World distibution} & Off &0.80 &0.91	&0.94	&0.99	&0.98\\
\hline
\textit{Baseline} & On & 0.53 & 0.45 & 0.60 & 0.93 & 0.96 \\
\textit{Location invariance} & On & 0.65 & 0.99 & 0.91 & 0.33 & 0.34 \\ 
\textit{Colour constancy} & On & 0.36 & 0.67 & 0.60	& 0.99 & 1.00 \\ 
\textit{World distibution} & On & 0.68 & 0.73 & 0.88 & 0.97 & 0.97 \\
\hline
\textit{Chance} &  & 0.33 & 0.33	 & 0.50 & 0.33 & 0.33 \\
\hline
\end{tabular}
\caption{Test accuracy of the five diagnostic classifiers for the four different games (average of three models). }
\label{tab:rnn}
\end{table*}

\paragraph{Perplexity per symbol}
The colour constancy game achieved the lowest perplexity per symbol, both with and without the vocabulary penalty.
This means that, on average, 1.3 and 2.2 symbols (respectively) were used to denote the same object, which is preferred over having many redundant symbols referring to the same object.

\paragraph{RSA values}
Even more revealing is the similarity between the representation of the objects in the agents' embedding spaces, which is what RSA depicts.
There is a high RSA Sender-Receiver score in all game variants, with scores peaking when the vocabulary penalty was applied.
High RSA Sender-Receiver scores are to be expected since a match on the embedding spaces of the agents is necessary for communication success.
However, it is the RSA with respect to the input that indicates whether the semantics of the agents' messages reflects the input structure.
Here, similarly to the perplexity per symbol metric, the colour constancy condition triggered the highest scores both for the Sender and the Receiver when the penalty is on. On the other hand, in absence of penalty, the location invariance game obtained the highest (absolute) RSA scores.

\paragraph{Topographic similarity}
A further indication that the location invariance condition has a positive effect on the semantics of the messages comes from topographic similarity: irrespective of the presence of the penalty, the highest score (i.e., the highest correlation between messages and the object space) was obtained in this game variant.\footnote{In the Appendix, we plot the development of agent-input RSA and topographic similarity across the training progress in the four games.}

\paragraph{Language entropy}
The location invariance game, with and without penalty, also achieved the lowest language entropy as it uses the least symbols of the vocabulary.

\subsection{Diagnostic classification of properties}
\label{subsec:diagnozze}

To inspect which properties of the input space are retained by the agent messages, we perform an analysis based on diagnostic classification \cite{hupkes2018visualisation}.
We train an RNN to encode the messages generated by the Sender and predict from its final hidden state the value for each symbolic property of the input image (shape, colour, size, horizontal position, vertical position).
Table \ref{tab:rnn} shows the accuracy of each classifier on the test messages.
The baseline model has the lowest scores for shape and colour, and is able to solve the task by mostly communicating row and column information.
On the other hand, the location invariance experiment cannot rely on position information, thus performing at a chance level as expected.
Rather, this model mostly encodes information about colour and size while playing the game, thereby supporting the hypothesis that the right environmental pressure encourages the encoding of higher-level information. 
The colour constancy setting seems to have some moderate impact on the colour semantics encoded by the messages.
The best results come once more from the world distribution game: a non-uniform (Zipfian) distribution of the objects induces a language that encodes, with high accuracy, all different properties of the image.


%% file: conclusion.tex
\section{Conclusions}
\label{sec:conclusion}

While most artificial agents achieve communication success in referential games, the emerging protocols are far from natural. Therefore, we coax the agent languages into developing desirable properties through sources of pressure that are both effective and realistic in terms of human cognition. In particular, we encourage the agents to make the least effort (in terms of sentence length and active vocabulary) through an auxiliary loss. Moreover, inspired by principles of perceptual constancy and frequency, we introduce external pressure by manipulating the appearance and frequency distribution of objects within images. Firstly, we found that least effort reduces message redundancy without loss of communication accuracy. Secondly, according to a series of well established metrics, external pressures facilitate the emergence of communicative protocols with a higher degree of compositionality. Thirdly, some sources of pressure such as colour constancy increase the accuracy in zero-shot communication, hence leading to a better ability to generalise. Finally, we reveal through diagnostic classifiers that agents under external pressures retain high-level information (such as shape or color of objects) rather than local pixel features. 
In general, the sources of pressure we propose bring forth a series of advantages: 1) they encourage more natural communication protocols; 2) they mitigate the arbitrariness of hyper-parameter setting; 3) they are realistic and justified by general principles of human cognition. 
In the future, this could help shedding light on the evolution of natural languages.

%% file: acknowledgements.tex
\section*{Acknowledgments}

DH is funded by the Netherlands Organization for Scientific Research (NWO), through a Gravitation Grant 024.001.006 to the Language in Interaction Consortium.
EB is funded by the European Union's Horizon 2020 research and innovation program under the Marie Sklodowska-Curie grant agreement No 790369 (MAGIC).
EMP is supported by the ERC Consolidator Grant LEXICAL (no 648909).

%% file: appendix.tex
\appendix
\section{Full Game Description}
\label{ap:setup}

\subsection{Agents}

\paragraph{Sender}
The inputs of the Sender are the feature representation $f$ of the target image $t$ (which we refer to as $f_t$) and a special start token \textless$S$\textgreater. 
Starting from an initial hidden state $h_0^S$, which is obtained by linearly transforming $f_t$, at each decoding step $i$ the Sender generates a token $w_i$ by sampling from its output distribution, until the special end of sequence token \textless$S$\textgreater is generated or the maximum sequence length $L$ is reached.


\paragraph{Receiver}
The Receiver receives the message $m$ generated by the sender as input.
It encodes this message and then uses a transformation of its last hidden state $h_l^R$ to select an image from the four images it is presented with (the correct target and $K=3$ distractors $d$).

\subsection{Training Signal}
The communication loss of the system is defined by
$$\mathcal{L}_c=\sum_{k=1}^K[\textrm{max}(0, 1 - q(t) + q(d_k))]$$
where the score function $q(x)=f_x^T g(h_l^R)$, where $g(x)$ is a linear function. Communication succeeds when the target's score is higher than all the distractors' scores.

Additionally, for the penalty variant of the game, we compute a vocabulary loss $\mathcal{L}_v$ defined in \S~4.1 of the main paper. The total loss is computed by the weighted sum $$\mathcal{L}=\mathcal{L}_c + \lambda \mathcal{L}_v$$

\section{Metrics across Training}
\begin{figure*}[!ht]
\centering
\includegraphics[width=\textwidth]{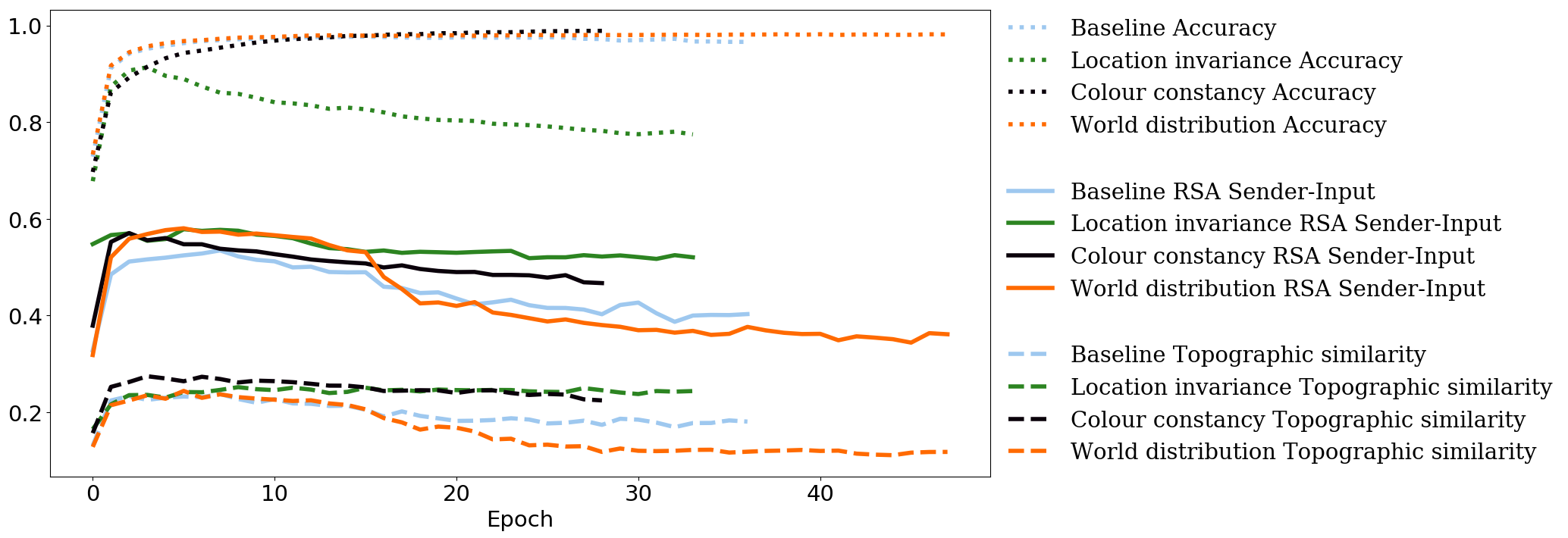}
\caption{Development of metrics with respect to the input for the four games during training with penalty.}
\label{fig:training_curves}
\end{figure*}

\begin{figure*}[!ht]
\centering
\includegraphics[width=\textwidth]{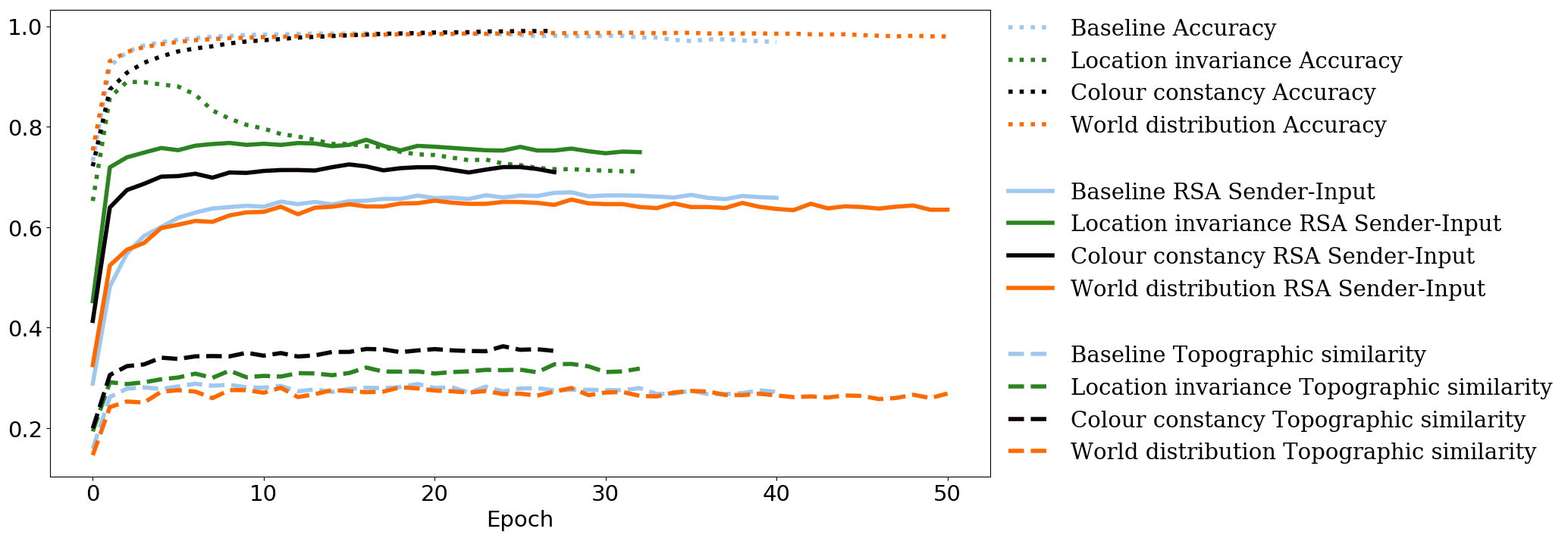}
\caption{Development of metrics with respect to the input for the four games during training without penalty.}
\label{fig:training_curves_no_penalty}
\end{figure*}

We provide information on the development of metrics such as accuracy, agent-input RSA and topographic similarity for the four games across training iterations. The plot for the setting with penalty is shown in Figure~\ref{fig:training_curves}, and for the setting without penalty in Figure~\ref{fig:training_curves_no_penalty}. These plots show how metrics can be damaged by overfitting for some game variants. On the other hand, some other variants preserve stable metric values across the entirety of the training iterations.

\section{Visual Module Training}
\label{ap:visual_acc}

The communication success obtained while playing the different games and training the corresponding CNN alongside is shown in Table \ref{tab:visual_module_acc}.

\begin{table*}[t]
\centering
\begin{tabular}{|c|c|c|c|} \hline
\textbf{Baseline} & \textbf{Location invariance} & \textbf{Colour constancy} & \textbf{World distribution} \\ \hline
0.97 & 0.89 & 0.89  & 0.98 \\ \hline
\end{tabular}
\caption{Test accuracy on the four different games when training the visual module.}
\label{tab:visual_module_acc}
\end{table*}

%

